\documentclass[lettersize,journal]{IEEEtran}
\usepackage{amsmath,amsfonts}
\usepackage{algorithmic}
\usepackage{algorithm}
\usepackage{array}
\usepackage[caption=false,font=normalsize,labelfont=sf,textfont=sf]{subfig}
\usepackage{textcomp}
\usepackage{stfloats}
\usepackage{url}
\usepackage{verbatim}
\usepackage{graphicx}
\usepackage{cite}
\usepackage{booktabs}
\usepackage{color}
\begin{document}

\title{Cycle-YOLO: A Efficient and Robust Framework for Pavement Damage Detection}


\author{Zhengji Li, Xi Xiao, Jiacheng Xie, Yuxiao Fan, Wentao Wang, Gang Chen, Liqiang Zhang, Tianyang Wang
\thanks{Zhengji Li is with College of Computer and software,Chengdu Jincheng College,Sichuan,China}
\thanks{Xi Xiao is with University of Alabama at Birmingham}
\thanks{Jiacheng Xie is with College of Computer and software,Chengdu Jincheng College,Sichuan,China}
\thanks{Yuxiao Fan is with College of Computer and software,Chengdu Jincheng College,Sichuan,China}
\thanks{Wentao Wang is with University of Alabama at Birmingham}
\thanks{Gang Chen is with Rovision Intelligent Technology Co,Sichuan,China}
\thanks{Liqiang Zhang is with College of Mathematics \& Computer Science,Guangxi Science \& Technology Normal University,Guangxi,China}
\thanks{Tianyang Wang is with University of Alabama at Birmingham}
}


\markboth{Journal of \LaTeX\ Class Files,~Vol.~14, No.~8, August~2021}%
{Shell \MakeLowercase{\textit{et al.}}: A Sample Article Using IEEEtran.cls for IEEE Journals}


\maketitle

\begin{abstract}
With the development of modern society, traffic volume continues to increase in most countries worldwide, leading to an increase in the rate of pavement damage Therefore, the real-time and highly accurate pavement damage detection and maintenance have become the current need. In this paper, an enhanced pavement damage detection method with CycleGAN and improved YOLOv5 algorithm is presented. We selected 7644 self-collected images of pavement damage samples as the initial dataset and augmented it by CycleGAN. Due to a substantial difference between the images generated by CycleGAN and real road images, we proposed a data enhancement method based on an improved Scharr filter, CycleGAN, and Laplacian pyramid. To improve the target recognition effect on a complex background and solve the problem that the spatial pyramid pooling-fast module in the YOLOv5 network cannot handle multiscale targets, we introduced the convolutional block attention module attention mechanism and proposed the atrous spatial pyramid pooling with squeeze-and-excitation structure. In addition, we optimized the loss function of YOLOv5 by replacing the CIoU with EIoU. The experimental results showed that our algorithm achieved a precision of 0.872, recall of 0.854, and mean average precision@0.5 of 0.882 in detecting three main types of pavement damage: cracks, potholes, and patching.On the GPU, its frames per second reached 68, meeting the requirements for real-time detection. Its overall performance even exceeded the current more advanced YOLOv7 and achieved good results in practical applications, providing a basis for decision-making in pavement damage detection and prevention.
\end{abstract}


\section{Introduction}
\IEEEPARstart{R}{oad} transportation is crucial for promoting communication between cities and countries. Moreover, road construction guarantees economic development, a prerequisite for people to travel safely, and is an indispensable condition for a country's economic development. \footnotetext{Abbreviations: KC: potholes, LF: cracks, XB: patches, ASPP: atrous spatial pyramid pooling, CBAM: Convolutional block attention module, FPS: frames per second, ReLU: rectified linear unit, SPPF: spatial pyramid pooling-fast, SSD: single-shot detector, mAP: mean average precision.}Cracks and damage are the most common problems encountered on the road surface\cite{ye2021convolutional, bock2021rubber, abdullah2021fatigue, hassan2022vulnerability}. \\

\begin{figure}[!ht]
    \centering
    \includegraphics[width=0.48\textwidth]{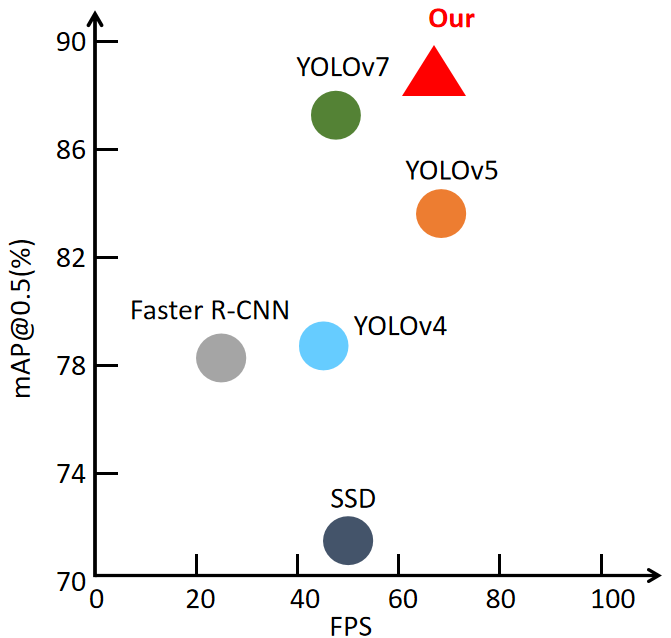}
    \caption{mAP and FPS comparison of 6 network models}
    \label{mAP and FPS comparison of 6 network models}
\end{figure}

Pavement damage detection is vital in road maintenance and road travel safety, as it can damage the pavement structure, reduce traffic speed, and shorten road operation time. Severe pavement damages can also weaken the bearing capacity of roadbeds, causing pavement collapse, traffic safety issues, and economic losses. Therefore, timely detection and effective maintenance of pavement damages are important for ensuring traffic safety, improving traffic efficiency, and maintaining the quality of travel. Nevertheless, accurately detecting traditional pavement damages in diverse and complex environments poses a significant challenge.At present, many scholars have proposed different pavement disease detection algorithms, and have obtained good results\cite{maeda2021generative, arya2021rdd2020, shim2022road, wang2022ensemble, jin2022road,gagliardi2022automatic, muttaqiy2024deteksi,palani2023road, mouzinho2021hierarchical, shim2021detection, dubeadvancing, bibi2021edge, ge2022tcnet, chen2022modality, zhao2021transformer3d, liu2021swinnet, yang2022bi, chen2021novel}. 
Currently, deep learning-based pavement damage detection methods are commonly used, including one-stage prediction methods based on regression classification, such as YOLO series\cite{wang2021improvements, wan2022yolo, he2023research,yang2024research,qian2023vehicle,huang2023lightweight,zhang2021vit,zhou2023intelligent,singh2024enhanced,kamalakannan2024novel,khan2024pothole,parmar2023drone,kumari2023yolov8,wei2023yolov5s,ruseruka2024augmenting,hong2024road,rout2023improved,du2022improvement,chen2024lag,fujii2023performance,yang4597443roadyolo,guo2022road,pham2022road,geng2023embedded,pham2023developing,okran2022effective}, and two-stage prediction methods based on region suggestion, such as the R-CNN series\cite{haciefendiouglu2022concrete,parvathavarthini2023road,naik2023pothole,lin2023research,kulambayev2023real,li2023damage,reddy2023deep,alsharafi2023investigation}. Although, the two-stage method has a high detection accuracy, its detection speed is poor. The one-stage method, which has a high detection speed while maintaining a certain detection accuracy, has gained importance in pavement damage detection. 

Cui et al.\cite{cui2021intelligent} built a dataset from concrete erosion tests and proposed an improved YOLO-v3 algorithm model to improve the detection capability of the model by modifying the loss and activation functions of the model and inserting a CSP module. PK et al.\cite{saha2024federated} deploys FL along with YOLOv5l to generate models for single- and multi-country applications and this research provides a strong foundation for further exploration of FL for road damage detection while preserving user privacy.Few studies have been conducted on the one-stage detection of pavement distress with a focus on the selection of models, dataset selection and processing, and the selection of loss functions. Nevertheless, these studies did not discuss the characteristics that distinguish pavement distress from general detection objects. The standard one-stage method models for pavement distress detection, such as the YOLO series, are designed to detect generic objects and do not fully consider the specific characteristics of pavement distress. Therefore, it is crucial to investigate how to improve them for pavement distress characteristics.

To address these challenges, the research team conducted an in-depth study and achieved the results\cite{li2022image}. Based on this, further research was conducted in this study to improve the CycleGAN and YOLOv5 algorithms, and a new pavement damage detection method was proposed by combining these two algorithms. In the experimental phase, we used YOLOv5, YOLOv7, and Faster R-CNN algorithms to test the proposed method on the collected dataset.

First, for the dataset category imbalance problem, we propose a data enhancement method based on an improved Scharr filter, CycleGAN, and Laplacian pyramid. Second, to address the shortcomings of the backbone network in damage feature extraction, we introduce a convolutional block attention module (CBAM) attention mechanism to enhance the feature extraction of the model for the region of interest. In addition, to extract multiscale features, increase useful information, suppress redundant information, and ensure the relevance of information acquired at a distance while maintaining high image resolution, we propose an atrous spatial pyramid pooling with squeeze-and-excitation (AS-SE) structure and replace the SPP module in the original YOLOv5 structure with it. By optimizing the loss function to reduce the loss value of the network, we achieved accelerated convergence speed of the network, thereby improving the accuracy of the entire detection network and realizing high-precision automatic detection and classification of pavement disorders.

In this study, our team spent three months collecting and collating 12,658 road image datasets to detect three main categories of road damages: potholes (KC), cracks (LF), and patches (XB). After cleaning considerable invalid data, we selected 1620 potholes (KC), 1230 cracks (LF), 794 repaired (XB) damage, and 4000 damage-free images for this study. However, the dataset suffers from class imbalance, some damage features are not clearly characterized, and the background environment is homogeneous, resulting in low detection accuracy, weak generalization, and poor robustness of the trained network models. To address these limitations, an enhanced pavement damage detection method with CycleGAN and YOLOv5 algorithm is proposed in this study.

\section{Method}
\label{Method}
\subsection{Overall algorithmic framework process}

\begin{figure*}[!htbp]
    \centering
    \includegraphics[width=\textwidth]{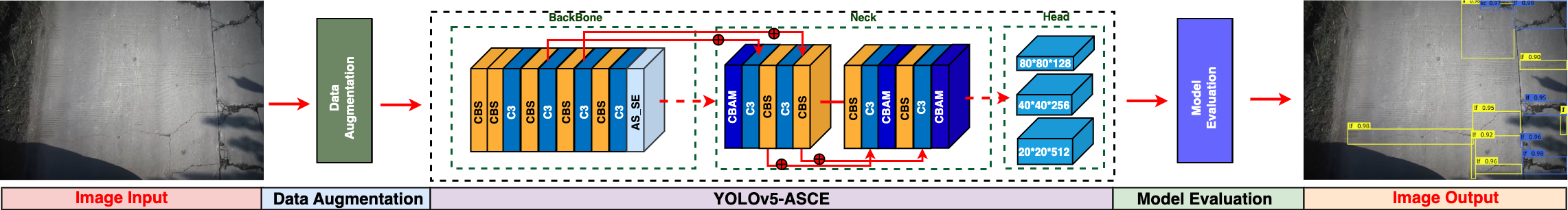}
    \caption{Overall Process Framework}
    \label{Overall Process Framework}
\end{figure*}

The algorithm used in this study is illustrated in Fig.2: First, the input images were data-enhanced, and then the data-enhanced image data were fed into the proposed YOLOv5-ASCE network. Figure 2 shows that we improved the feature extraction and fusion parts of the YOLOv5 network to make the final model more robust and general. Finally, the model was evaluated, and the detection images were output.

\subsection{Data enhancement methods based on improved Scharr filters, CycleGAN and Laplacian pyramid}
To improve the recognition accuracy of the target detection algorithms, data enhancement algorithms are typically used to enhance the datasets before training and recognition. Traditional data enhancement methods include the undersampling method, oversampling method, and image transformation\cite{zhu2017imbalance,batista2004study,elreedy2019comprehensive}, which can adjust the category distribution among samples to some extent but cannot account for the overall distribution characteristics of samples. These methods are not sufficiently diverse, and the final classification accuracy improvement is limited\cite{li2019review}. Meanwhile, more advanced image enhancement methods, such as TrivialAugment and AutoAugment\cite{müller2021trivialaugment,cubuk2019autoaugment}, suffer from a lack of variety, complexity, and adaptability to specific tasks.The cyclic consistent generative adversarial model (CycleGAN)\cite{zhu2017unpaired} learns the time-frequency features of the original data through adversarial games between generators and discriminators and controls the one-to-one mapping relationship between local data with the help of loss functions of multiple links to reconstruct the original data. The network can directly convert the original signal-time spectrogram to the generated signal-time spectrogram, and the generation process introduces an adversarial loss function and cyclic consistency loss function to ensure high-quality data generation.
\\\indent 
In this paper there are three types of image data such as cracks, potholes, and patches, and the use of CycleGAN allows for the generation of image variants that are not commonly found or covered in the existing data, such as detection targets of different sizes, shapes, or severities. This can further enhance the diversity of data to train more robust models. Also for special road damage situations (e.g., large cracks or pothole data) that are rare in naturally collected datasets, CycleGAN can be used to supplement the images with similar extreme cases, thus helping the model to learn to recognize and deal with such infrequent but important situations.

\begin{figure}[H]
    \centering
    \includegraphics[width=0.42\textwidth]{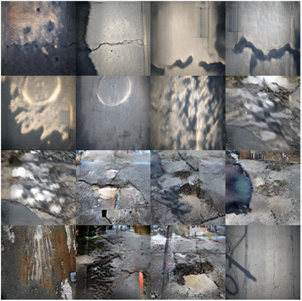}
    \caption{Non-real data images generated by CycleGAN}
    \label{Non-real data images generated by CycleGAN}
\end{figure}

However, it was observed that if the target damage images are generated directly by the CycleGAN network, the CycleGAN network is more sensitive to the input data and cannot handle complex scenes. When the number of features in the image being learned or the background is too complex, the CycleGAN network will learn various abstract features. For example, the combination of pavement distress with irrelevant features such as light shadows and the surrounding environment will produce some non-real data features, not only the distress features needed for this study, which leads to a large difference between the generated image and the real-world pavement image (as shown in Fig.3). This further leads to the deviation of the network model trained by the subsequent target detection neural network and its inability to achieve accurate detection of pavement damage.

As shown in Fig.3, the damaged pavement images generated directly using CycleGAN exhibited more interference terms, which were different from the real pavement images and did not satisfy the requirements of this study.
\\\indent Therefore, we propose a data augmentation algorithm based on CycleGAN and Laplacian pyramid. The data augmentation algorithm is illustrated in Fig.4.

\begin{figure}[H]
    \centering
    \includegraphics[width=0.48\textwidth]{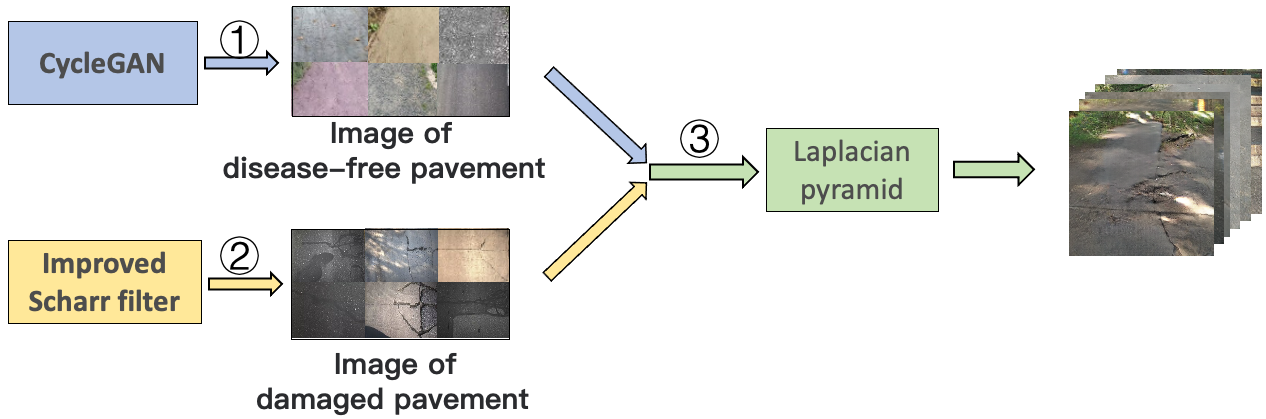}
    \caption{Data enhancement framework diagram}
    \label{Data enhancement framework diagram}
\end{figure}

In the first step, the CycleGAN network is used to generate intact pavement image data. In the second step, the images with obvious damage edge features in the dataset are filtered out by solving the gradients in the 0°, 45°, 90°, and 135° directions of the images, that is, the image data with more obvious damage features in the training set. After completing the aforementioned two steps, the image generated by CycleGAN in the first step is used as the background image, the damage feature area obtained from the gradient information in the second step is used as the foreground image, and the two are fused by a Laplacian pyramid to obtain the image data after data enhancement.

\subsubsection{Image gradient solving}
To filter image data with obvious damage features more effectively, the Scharr filter\cite{weickert2002scheme} was used to solve the image gradient of the original image. The Scharr filter has the advantages of fast computation and high accuracy and can extract weak edge features while being sensitive to the influence of adjacent pixels and grayscale changes. However, as shown in Fig.5, the conventional Scharr filter computes the gradient only for the 0° and 90° directions of the image and does not consider the gradient in the diagonal direction, which is an essential source of information for the edge and texture features in image processing.

\begin{figure}[H]
    \centering
    \includegraphics[width=0.2\textwidth]{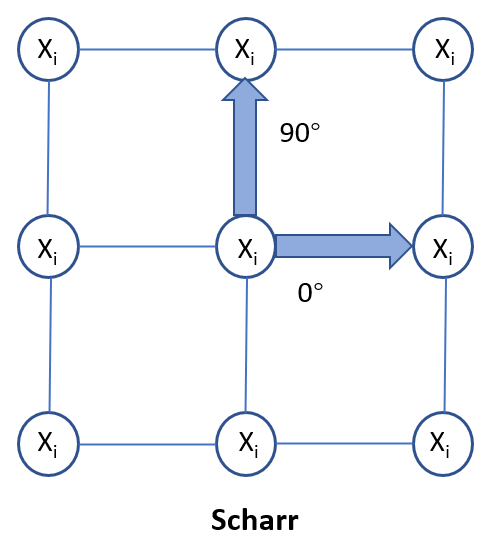}
    \caption{Scharr filter schematic}
    \label{Scharr filter schematic}
\end{figure}

Solving the gradient in only two directions (e.g., 0° and 90°) might miss some features in the diagonal direction, whereas solving the gradient in four directions (0°, 45°, 90°, and 135°) can extract these important features more comprehensively and help improve the accuracy of image analysis. Moreover, compared to solving the gradient in only two directions, solving the gradient in four directions can better preserve the detailed information of the image and thus improve its resolution. Therefore, to improve the accuracy of the gradient approximation and increase the resolution of the image, an improved Scharr filter is proposed to calculate the gradients of the image in four directions: 0°, 45°, 90°, and 135°. The improvement principle is shown in Fig.6, where Xi (i=1,2,3,..., n) denotes each pixel point in the image, and the gradient calculation in the four directions can be used to analyze the entire image more comprehensively.

\begin{figure}[H]
    \centering
    \includegraphics[width=0.2\textwidth]{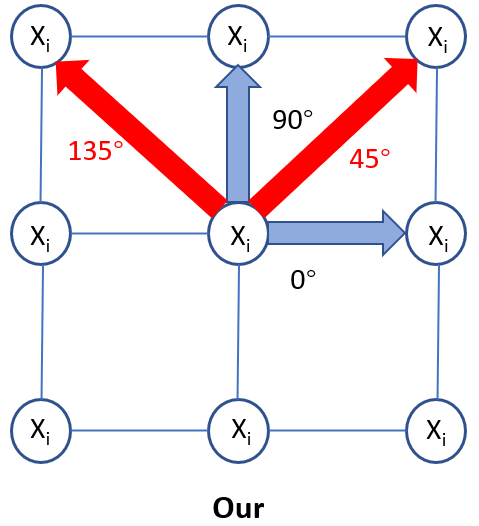}
    \caption{Improved Scharr filter schematic}
    \label{Improved Scharr filter schematic}
\end{figure}

Specifically, the proposed modified Scharr filter solves the gradient in four directions (0°, 45°, 90°, and 135°) and is significantly advantageous over the original Scharr filter, which solves the gradient using only two directions (0° and 90°).
\\\indent To verify the effectiveness of the improved Scharr filter, a comparison experiment with a gradient solution was conducted on the same pavement damage image. In the experiments, the gradient solutions of the images were obtained using the original and improved Scharr filters and the gradient images of the damage features were obtained, as shown in Fig.7. The experimental maps were divided into three groups––A, B, and C––for comparison at various magnifications. Specifically, Group A shows a comparison of feature images at 100\% magnification, Group B shows a comparison of feature images at 300\% magnification, and Group C shows a comparison of feature images at 500\% magnification. By carefully observing and comparing the experimental results, we can observe that the feature gradients solved with the improved Scharr filter performed better in terms of clarity and distinctness than the feature images obtained with the original Scharr filter. This finding illustrates the significant advantage of the improved Scharr filter in extracting image details and highlighting damage features. In addition, it provides strong support for research on damage detection and identification.
\\\indent Therefore, it can be concluded that the proposed Scharr filter can extract the features of pavement damages more effectively, thereby laying a solid foundation for damage identification.

\begin{figure}[H]
    \centering
    \includegraphics[width=0.48\textwidth]{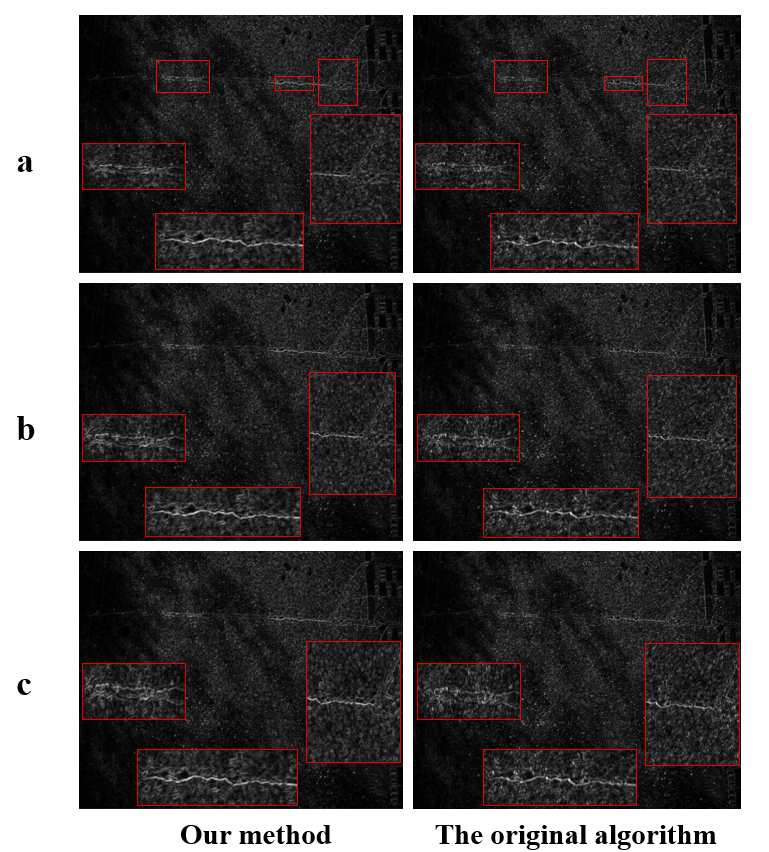}
    \caption{Comparison of the solution results}
    \label{Comparison of the solution results}
\end{figure}

\subsubsection{Image fusion}
To obtain high-quality pavement damage image data, we initially tried to use the image stitching method to generate the required image data. However, in practice, the image data obtained by this method had apparent edge seams at the stitched edge and differences in color and brightness between the before- and after-background images.
\\\indent To solve this problem, we used Gaussian blur to process the mask region so that the junction between the foreground and background interpenetrates, the grayscale mutation is smoothed, and the foreground and background images are blended. Although more natural image data can be obtained after this process, the computational effort increases significantly with an increase in the Gaussian blur radius and an appropriate Gaussian blur radius cannot be determined. In addition, the limited resolution of the images generated using CycleGAN makes the entire process time-consuming and has a limited effect.
\\\indent To acquire the simulation data images more efficiently and solve the aforementioned problems, we improved the image fusion method using the Laplacian pyramid\cite{burt1987laplacian} for image fusion. The Laplacian pyramid method can smooth the edges between the foreground and background better while maintaining a low computational effort. This method can overcome the limitations of previous methods and achieve higher quality and natural-feeling image data fusion, thus providing more effective simulation data for pavement damage image studies. One of the roles of the Laplacian pyramid is to recover high-resolution images (see Fig.8 for details).
\\\indent The Laplacian Pyramid is defined by: Li = Gi - $pyrUp$(Gi + 1)
\\\indent Here, Li denotes the \textit{i}th level in the Laplacian pyramid, Gi denotes the \textit{i}th level in the Gaussian pyramid, and Gi+1 denotes the \textit{i}+1th level in the Gaussian pyramid.

\begin{figure}[H]
    \centering
    \includegraphics[width=0.48\textwidth]{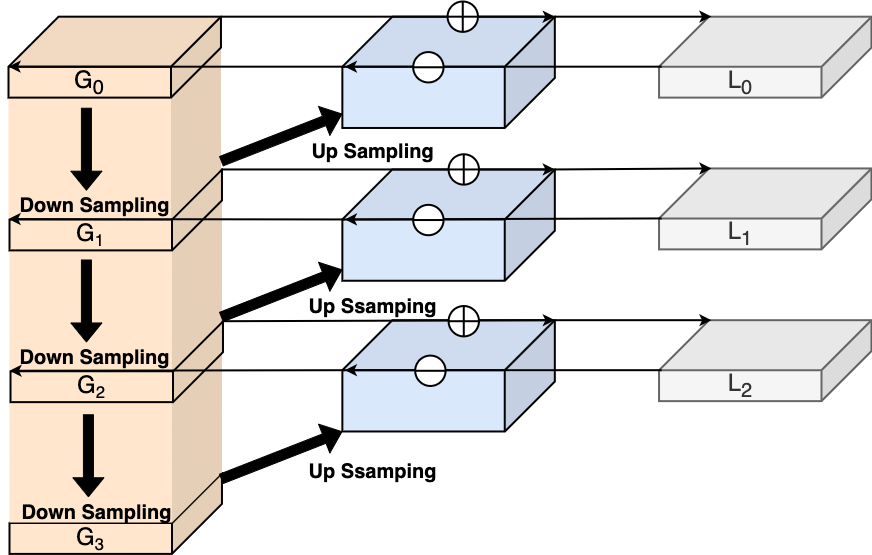}
    \caption{Process chart of Laplacian Pyramid to recover high-resolution images}
    \label{Process chart of Laplacian Pyramid to recover high-resolution images}
\end{figure}

\begin{figure*}[!htbp]
    \centering
    \includegraphics[width=0.8\textwidth]{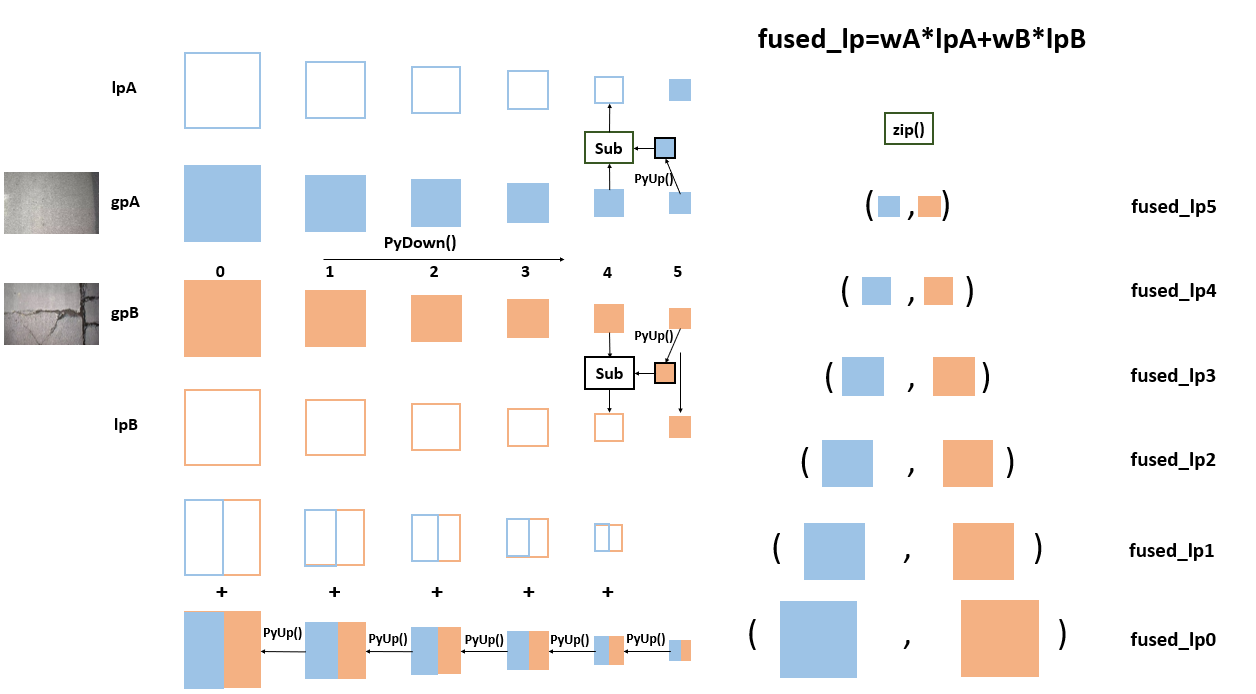}
    \caption{Laplacian pyramid image fusion process chart}
    \label{Laplacian pyramid image fusion process chart}
\end{figure*}

The meaning of each marker in the figure above is as follows: G0, G1, G2, and G3 are levels 0, 1, 2, and 3 of the Gaussian pyramid, respectively. L0, L1, and L2 are levels 0, 1, and 2 of the Laplacian pyramid, respectively. The downward arrow indicates the downsampling procedure. The arrow in the upper right indicates the upsampling operation. The “+” sign indicates an additional operation. The negative sign "-" indicates a subtraction operation.
\\\indent The core principle of Laplacian pyramid image fusion is as follows:
\\\indent (1) Create a Gaussian pyramid
\\\indent a. For image A, the $PyDown()$ function was used to create a Gaussian pyramid(gpA). The $PyDown()$ function generates a series of images with a progressively lower resolution by downsampling and blurring the original images.
\\\indent b. For image B, the $PyDown()$ function was used to create a Gaussian pyramid (gpB). Processed in the same manner as in Image A.
\\\indent (2) Creation of the Laplacian pyramid
\\\indent a. Create the corresponding Laplacian pyramid(lpA) based on the Gaussian pyramid gpA of image A. Use the suboperation (subtraction operation) and the $PyUp()$ functions. For each layer, the image of the next layer was subtracted from the image of the current layer of the Gaussian pyramid (after upsampling and blurring using the $PyUp()$ function).
\\\indent b. Create the corresponding Laplacian pyramid (lpB) based on the Gaussian pyramid gpB of image B. Process in the same manner as for image A.
\\\indent (3) Create weight images
\\\indent Generate weighted images that determine the weights of images A and B at each pixel position. This can be achieved using features such as the gradient and brightness of the images. Suppose that the weight images are wA and wB.
\\\indent (4) Fusion Laplacian pyramid
\\\indent For each input image, Laplacian pyramids (lpA and lpB) and weighted images (wA and wB) were used to determine the weight of each pixel in the final composite image. The weighted Laplacian pyramid was summed by elements to obtain the fused Laplacian pyramid (fused-LP). 
\\\indent the specific operation is as follows:$fused\_lp = wA * lpA + wB * lpB$
\\\indent (5) Reconstructing the fused image
\\\indent The fused-LP was processed layer-by-layer using the zip() function. Starting from the coarsest layer, the image of the current layer is upsampled and blurred using the $PyUp()$ function and then added to the image of the next layer (using the Sub operation) until the finest layer is obtained. The final fused image is shown in Fig.9. The pavement damage image data obtained using this method are closer  to the actual pavement damage image, and the operation is convenient and effective, which is beneficial for our research.

\subsubsection{CycleGAN network}
In this study, we propose a CycleGAN network-based data enhancement approach for generating diverse and complex backgrounds of damage-free original pavement images and fusing the damage features in the filtered damage-featured images with the generated intact pavement images to obtain substantial pavement image data with different backgrounds and distinct damage features, which is convenient for our subsequent research.
\\\indent 
CycleGAN is an unsupervised machine learning method developed from a generative adversarial networkGAN,  which is an unpaired image-to-image transformation network\cite{goodfellow2020generative}. The purpose of CycleGAN is to achieve the interconversion of X-domain image data and Y-domain image data, which contains two mapping functions $G: X\rightarrow Y$,$F:Y\rightarrow X$, and two discriminators D\_X, D\_Y. For generator G, the X-domain can be transformed to Y-domain data. For generator F, the reverse transformation of the Y-domain to X-domain data. Discriminator D\_X motivates generator F to shift Y to X. Similarly, discriminator D\_Y motivates generator G to shift X to Y. The conversion relationship is illustrated in Fig.10.
\\\indent 
Two adversarial losses were introduced into the CycleGAN and applied to the two mapping functions. For the mapping function $G: X\rightarrow Y$, a given sample $x \in X$ is transformed into data in domain Y with discriminator D\_Y using the following relationship:

\begin{equation}
    \begin{split}
        L_{G A N}\left(G, D_{Y}, X, Y\right) & =E_{y \sim p_{\text {data }}(y)}\left[\log D_{Y}(y)\right]  \\
        & + E_{x \sim p_{\text {data }}(x)}\left[\log \left(1-D_{Y}(G(x))\right]\right. 
    \end{split}
\end{equation}

In Eq (1), $p_{data} (x)$ represents the X-domain sample space distribution; $p_{data} (y)$ represents the Y-sample space distribution; the generator G tries to generate images as similar as possible to those in the Y-domain, while $D_Y$ tries to distinguish the samples from the real picture y. The goal of G is to minimize the image, whereas the goal of $D_Y$ is to maximize, and the two form an adversarial relationship, $_{min}$G$_{max}$$D_Y L_{GAN} (G,D_Y,X,Y)$. Similarly, there exists a mapping function$F:Y\rightarrow X$, which is related to discriminator $D_X$ as follows:

\begin{equation}
    \begin{split}
        L_{G A N}\left(F, D_{x}, Y, X\right) & =E_{x \sim p_{\text {data }}(x)}\left[\log D_{X}(x)\right] \\
        & + E_{y \sim p_{\text {data }}(y)}\left[1-\log D_{X}(F(y))\right]
    \end{split}
\end{equation}

Similarly, the objective of F is to minimize the image, whereas the objective of D is to maximize it, and there is an adversarial relationship, $_{min}$F$_{max}$$D_X L_{GAN} (G,D_Y,Y,X)$.

Theoretically, adversarial training can learn mapping G and F to produce outputs with the same distribution as the target domains Y and X, respectively. However, adversarial loss alone may not be able to map the same set of input images to arbitrarily randomly arranged images in the target domain, which is prone to loss of image information. CycleGAN addresses this limitation by introducing cyclic consistency loss, as shown in Fig.10-(b) and Fig.10-(c), and for each image x in the X domain, x can be reduced to the original image by cyclic transformation, i.e., positive cyclic consistency loss $x\rightarrow G(x)\rightarrow F(G(x))\approx x$. Similarly, for each image y in the Y domain, the reverse cyclic consistency $y\rightarrow F(y)\rightarrow G(F(y))\approx y$ should be satisfied.

\begin{figure}[H]
    \centering
    \includegraphics[width=0.48\textwidth]{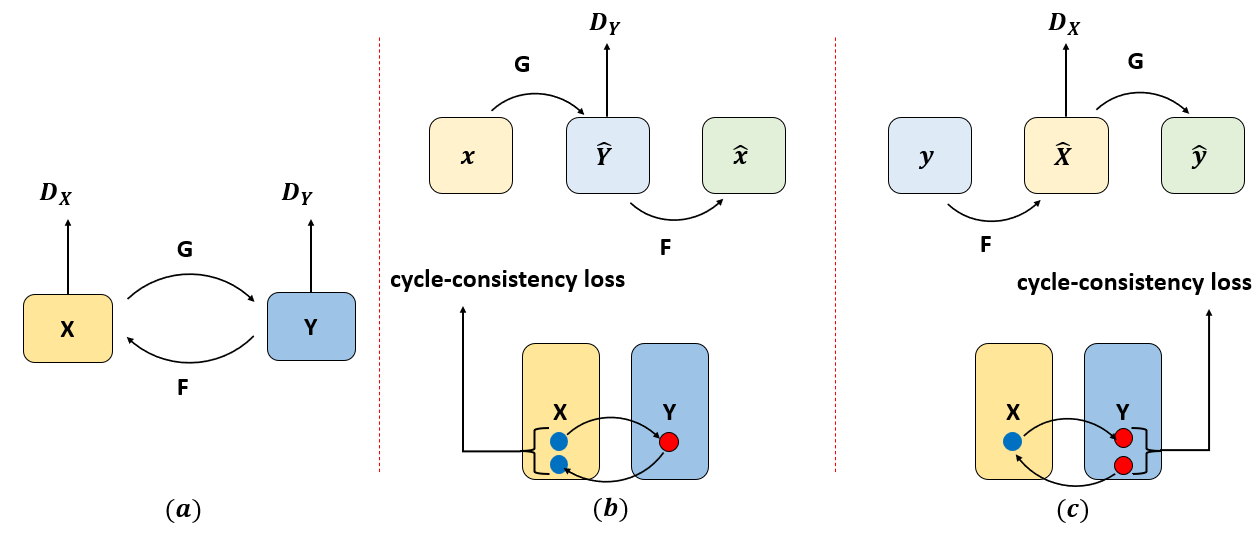}
    \caption{Structure diagram of cycle-consistent generative adversarial network}
    \label{Structure diagram of cycle-consistent generative adversarial network}
\end{figure}

Definable cyclic consistency loss:

\begin{equation}
    \begin{split}
        L_{cyc}(G,F)= & E_{x\sim p_{data}(x)}[||F\big(G(x)\big)-x||_1] \\ 
        & +E_{y\sim p_{data}(y)}[||G\big(F(x)\big)-y||_1]
    \end{split}
\end{equation}

In Eq (3): G is the forward generator; $F(y)$ is the reverse generator: $F(G(x))$ and $G(F(y))$ are the reconstructed images.

\begin{figure}[H]
    \centering
    \includegraphics[width=0.48\textwidth]{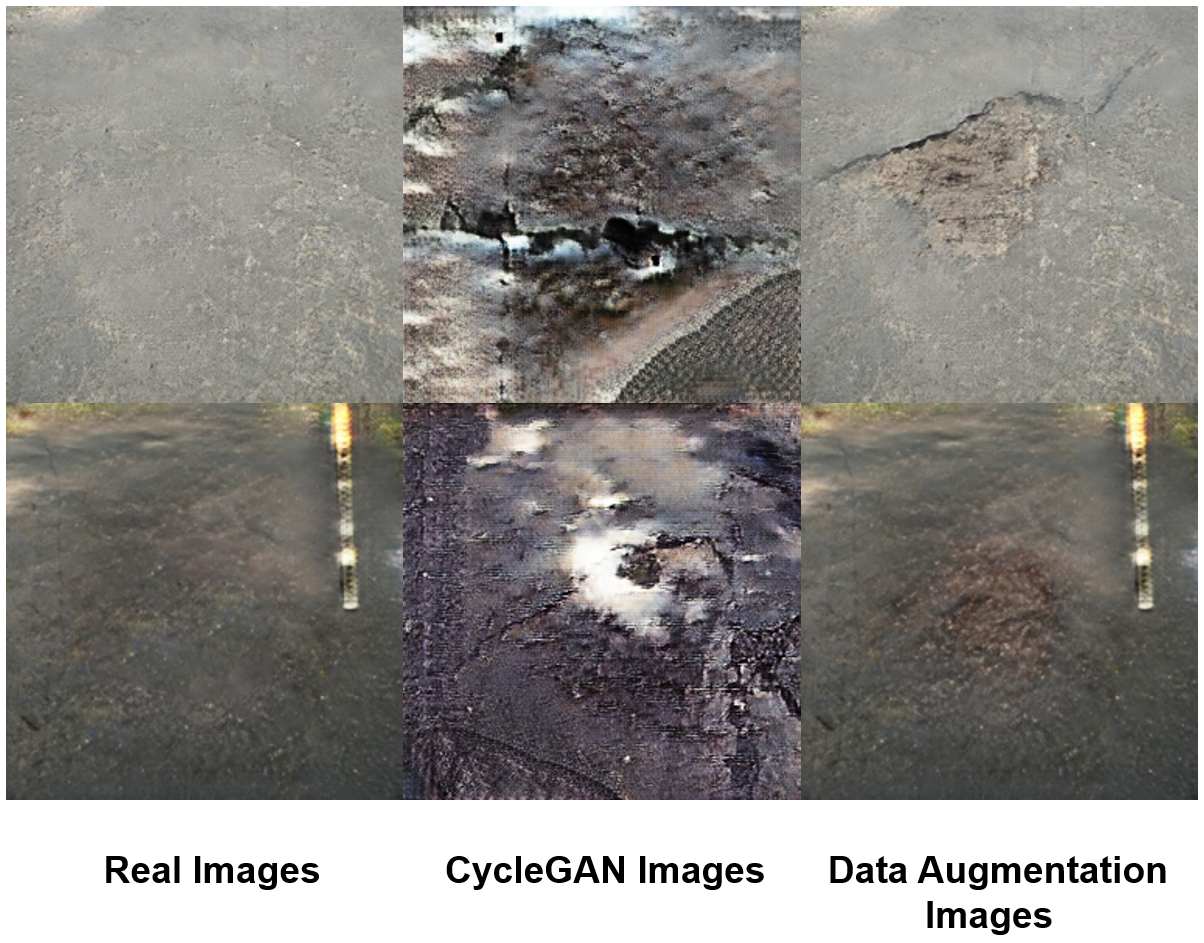}
    \caption{Comparison of real image with CycleGAN generated image and data enhanced image}
    \label{Comparison of real image with CycleGAN generated image and data enhanced image}
\end{figure}

Fig.11 shows that the damage data obtained by the proposed data enhancement process are closer to the real road surface compared with the damage data obtained by directly using the CycleGAN algorithm. The image data have no obvious edge seams, and the fused parts of the edges are smooth, which significantly improves the quality and size of the dataset. 
To verify the effectiveness of the data enhancement algorithm proposed in this paper, a test comparison of the dataset before and after data enhancement using the YOLOv5 algorithm was conducted in this study(The experimental platform and parameter settings are placed in section IV).
The results are shown in Table 1. It can be seen that the commonly used data enhancement methods do not work well for data processing for specific detection targets. While under the data enhancement method proposed in this paper, compared with the original dataset, the precision improved by 13.8\%, recall improved by 10.1\%, and mean average precision (mAP)@0.5 improved by 8.1\%.


\begin{table*}[!ht]\normalsize
\tabcolsep=1cm 
\caption{\textbf{Comparison of data enhancement}}
\begin{tabular*}{\linewidth}{@{}lccc@{}}
\toprule
\textbf{Method} & \textbf{Percision(\%)} & \textbf{Recall(\%)} & \textbf{mAP(\%)} \\
\midrule
Undersampling & 83.5 & 52.0 & 70.4 \\
Oversampling & 85.2 & 61.2 & 75.11 \\
AutoAugment & 79.6 & 39.4 & 53.7 \\
Trivialaugment & \textcolor{blue}{85.9} & 46.0 & 62.8 \\
Original Dataset & 77.8 & \textcolor{blue}{68.7} & \textcolor{blue}{77.0} \\
\textbf{Our algorithm} & \textbf{87.1} & \textbf{81.9} & \textbf{85.0} \\
\bottomrule
\end{tabular*}
\end{table*}



The CycleGAN network used in this study was only used to generate intact pavement background image data, which were then fused with the distinctive damage dataset filtered by the Scharr filter using a Laplacian pyramid to generate a complete road damage dataset. This method can generate numerous real pavement damage datasets with different backgrounds and distinctive features. This dataset is highly simulated and usable and lays a solid foundation for subsequent research and experiments.\\

\section{Improved YOLOv5 network structure}
After completing the data enhancement part, we analyzed the YOLOv5 network structure and found that its spatial pyramid pooling-fast (SPPF)\cite{he2015spatial} module used as a spatial pooling operation affected the image resolution, which led to the loss of detailed information and affected the model detection effect. Thus, this module was replaced with the ASPP (atrous spatial pyramid pooling)\cite{chen2017deeplab} and the AS-SE structure proposed by squeeze-and-excitation networks (SENets)\cite{hu2018squeeze} to ensure the high resolution of the image while expanding the perceptual field during the spatial pooling operation and to make the channel features of the image more obvious. To improve the efficiency of the feature fusion network, we introduce CBAM\cite{woo2018cbam} in the three feature fusion layers at different scales of the YOLOv5 feature fusion network, which enhances the saliency of the target in the complex background, thus improving the feature expression of the foreign object target and the detection accuracy of the detection network. In response to the aforementioned problems and improvement ideas, the YOLO-ASCE network structure was proposed in this study, as shown in Fig.12, to achieve the accurate and fast detection of pavement damage targets.

\begin{figure}[H]
    \centering
    \includegraphics[width=0.48\textwidth]{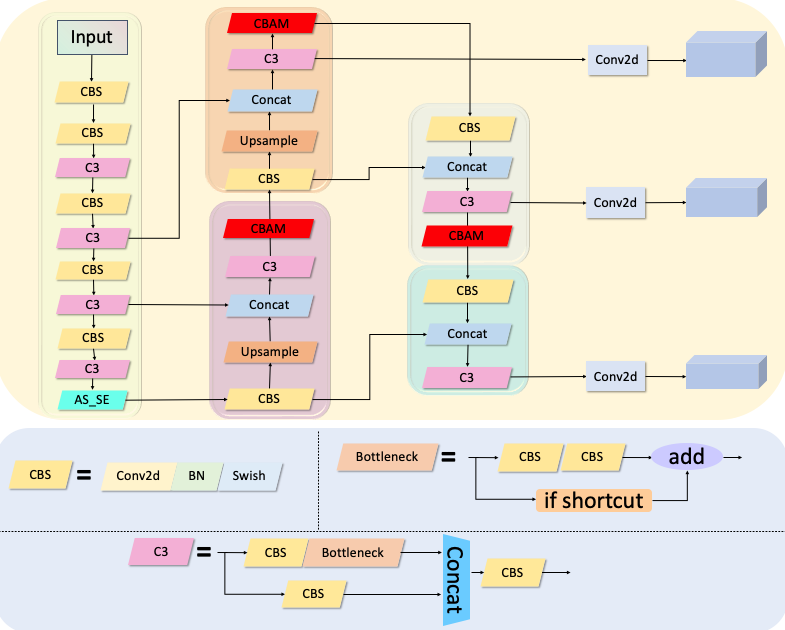}
    \caption{YOLO-ASCE network structure}
    \label{YOLO-ASCE network structure}
\end{figure}

\subsection{CBAM(Convolutional Block Attention Module)}

\begin{figure*}[!htbp]
    \centering
    \includegraphics[width=\textwidth]{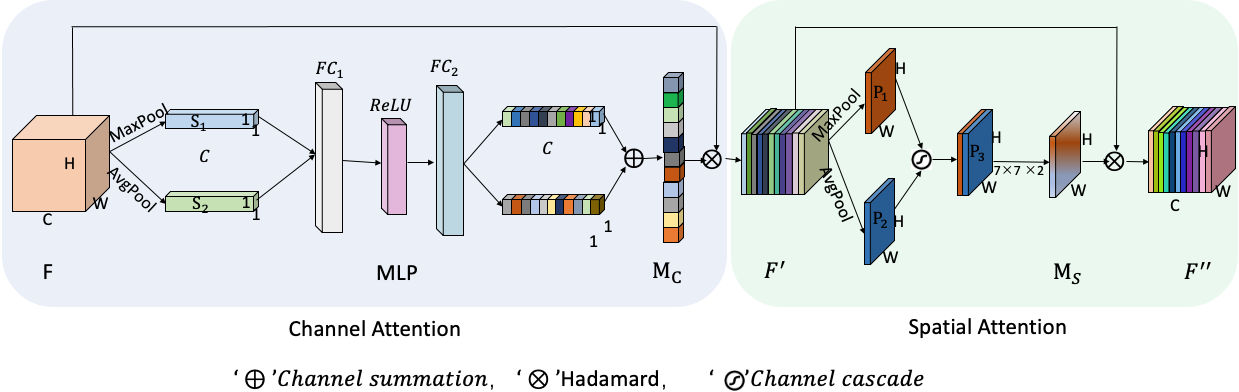}
    \caption{CBAM structure}
    \label{CBAM structure}
\end{figure*}

The main purpose of integrating CBAM in the YOLO framework is to optimize the model's focus on important features in the image. For pavement damage data where there may be multiple areas with fragmented distribution of features, enhancing these features with CBAM helps the model to focus more on areas where damage may be present. For this reason this study introduces the CBAM module after each C3 module in the downsampling phase, i.e., it is inserted before feature fusion. It also ensures that the model prioritizes learning from the parts of the image that are most likely to contain cracks, potholes, or repairs, thus improving detection accuracy.
\\\indent 
The CBAM module was added to the feature fusion layer at three different scales to improve the efficiency and model detection performance of the feature fusion network. The overall process of the CBAM attention mechanism is shown in Fig.13. It comprises a channel attention mechanism and a spatial attention mechanism to collaboratively learn the key local detail information in the image, improve the attention of the neural network to the damage in the image, and improve the network's feature learning and expression capability. This module reduces the computational overhead required by the convolutional multiplication method, reducing its module low in complexity and computation. Thus, it was added to optimize the network structure to achieve faster and more accurate capture of the region of interest.

\subsubsection{Channel attention mechanism}
Each channel of the feature map has a different level of importance for determining whether a pixel point belongs to a road defect. Therefore, a channel attention mechanism is introduced in this study to establish the interrelationship between the channels of the feature map such that the feature information of different channels can be used more effectively.
\\\indent 
As shown in the left channel attention section in Fig.13, the input feature map is denoted as $F \in R^{(C\times H\times W)}$, H and W are the height and width of the input feature map, respectively, and C is the number of channels of the input feature map. The global spatial information of feature map F is compressed by global Mmaximum Ppooling and global Aaverage Ppooling to generate two feature maps, $S_1$ and $S_2$ of size $1\times 1\times C$. The position information of each channel feature map in the entire map was fused to avoid network evaluation bias caused by the small size of the convolutional kernel when extracting features, owing to the small perceptual field range.
\\\indent 
$S_1$ and $S_2$ obtain two one-dimensional feature maps by sharing a multilayer perceptiron (MLP)  consisting of fully connected layers $FC_1$, $FC_2$ and rectified linear unit ( ReLU) nonlinear activation functions. The two one-dimensional feature maps are normalized by the sigmoid function after the summation operation by channel to obtain the weight statistics $M_C$ of size $1\times 1\times C$ for each channel, which can be expressed by Eq (4):

\begin{equation}
    \begin{split}
        M_C(F)& =\sigma\{MLP[AvgPool(F)]+MLP[MaxPool(F)]\}  \\
        &=\sigma\{MLP[\frac{1}{H\times W}\sum_{i_0=1}^H\sum_{j_0=1}^W f_{x_0}(i_0,j_0)]+ \\
        & MLP[\operatorname*{max}_{i\in Hj\in W}f_{x_0}(i_0,j_0)]\}
    \end{split}
\end{equation}

In Eq (4): $\sigma$ denotes the sigmoid function, $f_{x0}(i_0,j_0)$ denotes the pixel value of the coordinate position $(i_0,j_0)$ point in the $x_0$th channel of the input feature map F.

\subsubsection{Spatial attention mechanism}
As shown in the right spatial attention part in Fig.13, the spatial attention mechanism takes the feature map $F'$ obtained after channel feature rescaling as input, and does global maximum pooling and average pooling operations in channel dimensions to obtain feature maps $P_1\in R^{(1\times H\times W)}$ and $P_2\in R^{(1\times H\times W)}$, respectively. The feature description $P_3\in R^{(2\times H\times W)}$ is obtained by channel cascading, and the information at different locations in $P_3$ is encoded and fused using convolutional layers to obtain spatially weighted information $M_{S}$, which can be used to distinguish the importance of different spatial locations of the image, and the process can be expressed by Eq (5):

\begin{equation}
    \setlength{\abovedisplayskip}{-0.45cm}
    \setlength{\abovedisplayskip}{-0.05cm}
    \begin{split}
        M_S(F)=\sigma\{f^{7\times7}[AvgPool(F');MaxPool(F')]\}
    \end{split}
\end{equation}

In the Eq (5): $f^{7\times7}$ indicates that the convolution kernel is a convolution layer of size 7*7.

\subsubsection{CBAM attention mechanism}
The input feature map F is multiplied with the corresponding elements of the two matrices of each channel weight value $M_C$, and the feature rescaling process is performed on F to obtain the feature map $F'$ that can effectively reflect the feature key channel information. Based on channel weighting, the spatial feature information is adaptively weighted using the tandem spatial attention mechanism, and $F'$ is multiplied by the corresponding elements of the two matrices of spatial weight coefficients $M_S$ as the input of the spatial attention module to obtain the significant feature map $F''$ containing the channel location information and spatial location information. Thus, the network can enhance the attention to on the road diseasedamage input features that are strong and enhance its spatial feature selection capability.
\\\indent 
The process can be expressed as Eq (6):

\begin{equation}
    \begin{cases}
        F'=M_C\otimes F  \\ 
        F''=M_S\otimes F'
    \end{cases}
\end{equation}

In Eq (6): $'\otimes'$ means that the corresponding elements of the two matrices are multiplied together.

\subsection{AS-SE module}
\subsubsection{ASPP(Atrous Spatial Pyramid Pooling)}

The main purpose of integrating ASPP in the YOLO framework is to optimize the model's ability to handle the scale variability of pavement damage data. For pavement damage data with a large interval range of crack and pothole damage sizes and depths, the sensory field of the convolutional neural network can be expanded without affecting the resolution of the pavement damage images through ASPP,thus improving the detection accuracy of cracks,potholes and repairs.
\\\indent
The SPPF module in the YOLOv5 network structure affects the image resolution, which results in the loss of detailed information and affects the model detection effect, while the hole convolution reduces the dependence on parameters and computational process in the calculation on the basis of ensuring the image resolution property; fewer parameters are required to achieve the effect of expanding the effective sensory field of the convolution kernel, which can effectively aggregate the contextual information. A two-dimensional atrous convolution for each position i of the output feature map B and filter w applies atrous convolution to the input feature map x:

\begin{equation}
    \setlength{\abovedisplayskip}{-0.45cm}
    \setlength{\abovedisplayskip}{-0.05cm}
    \begin{split}
        B[i]=\sum_{k=1}^k A[i+r\cdot k]\cdot w[k]
    \end{split}
\end{equation}
In Eq (7), \textit{k} denotes the convolution kernel size, and \textit{r} denotes the sampling rate. Eq (3-4) indicates that \textit{r}-1 zero values are inserted into the new filter along each spatial dimension between two consecutive filters, and then the feature map A is convolved through this filter, and $B[i]$ represents the final obtained feature map. Thus, atrous convolution can control the perceptual field of the filter and the compactness of the network output features by changing the sampling rate without increasing the number of parameters or computational effort.
\\\indent 
The multiscale fusion of ASPP uses atrous convolution with a multilevel atrous sampling rate to sample feature maps in parallel, enabling the ASPP module to learn image features from different sensory fields. Because the atrous convolution with a large sampling rate degrades to a 1$\times$1 convolution owing to the inability of the image boundary response to capture long-range information, the image-level features obtained by global average pooling are fused into the ASPP module. That is, the image-level features and the feature map output from the four convolution branches are input into a 1$\times$1 convolution layer and then bilinearly upsampled to a specific spatial dimension, which is computed as in Eq (8).

\begin{equation}
    \setlength{\abovedisplayskip}{-0.45cm}
    \setlength{\abovedisplayskip}{-0.05cm}
    \begin{split}
        \text{B=Concat} & (image(A),H_{1,1}(A),H_{6,3}(A), \\
        & H_{12,3}(A),H_{18,3}(A))
    \end{split}
\end{equation}

In Eq (8), $H_{r,n}(A)$ represents the atrous convolution of A with sampling rate r and convolution kernel size $n\times n$; image(A) represents the extraction of image-level features from input A using the global average pooling method, and the ASPP structure is shown in Fig.14.

\begin{figure}[H]
    \centering
    \includegraphics[width=0.40\textwidth]{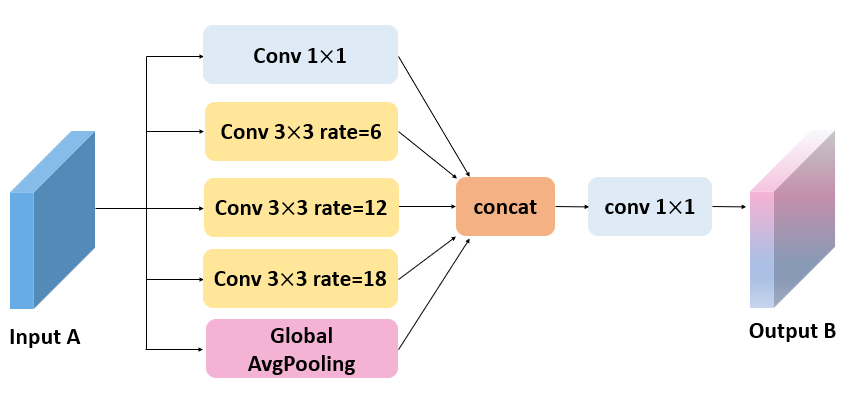}
    \caption{ASPP module structure}
    \label{ASPP module structure}
\end{figure}

Because the SPPF module in the YOLOv5 network structure is computationally intensive, the feature map information is limited, and the pooling operation loses some detailed information\cite{tan2020efficientdet}, while the above ASPP structure expands the perceptual field and enriches the semantic information by the multilevel sampling rate of atrous convolutional parallel sampling. Thus, the image-level features can effectively capture the global contextual information and consider the relationship between contexts, avoiding the problem of getting into local features that lead to segmentation errors, and improve the segmentation accuracy of the target. Therefore, the feature maps containing high-level semantic information are input to the ASPP module for obtaining features at different scales before upsampling, which helps improve the network's performance for road damage extraction.

\subsubsection{SENet(Sequeeze-and-Excitation)}

The main objective of integrating SENet in the YOLO framework is to enhance the feature recalibration capability between model channels. In the context of pavement damage detection, SENet helps the model to dynamically adjust the weights of each channel based on the relevance of a particular type of damage to the features, thus enhancing the model's ability to recognize more subtle features of pavement damage.
\\\indent
The core part of the SENet is the SE module that can be embedded in other classification or detection models. The structure comprises two parts, squeeze and excitation, which automatically obtain channel weights and capture spatial correlations through learning, thus improving the expressiveness of the network. The structure of the SENet module is illustrated in Fig.15.

\begin{figure}[H]
    \centering
    \includegraphics[width=0.48\textwidth]{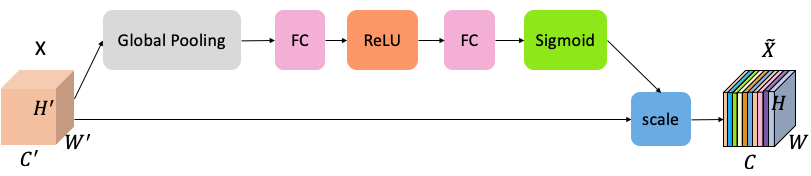}
    \caption{SE module structure}
    \label{SE module structure}
\end{figure}

First, the global information of $H\times W\times C$ is compressed into one channel of $1\times 1\times C$ by pooling through a compression process (squeeze), which widens the sensory field using a global average pooling algorithm. The equation used is as follows:

\begin{equation}
    \begin{split}
        F_{sq}(U_c){-}\frac{1}{W\times H}\sum_{i=1}^W\sum_{i=1}^H U_c(i,j)
    \end{split}
\end{equation}

The activation process (excitation) is then obtained. After further compressing and reconstructing the features by two fully connected layers (Fully-connected), the ReLU\cite{glorot2011deep} and sigmoid\cite{cybenko1989approximation} are connected to the activation layer to obtain the weight information of the feature dimensions, and the purpose is to capture the relationship between each channel.

\begin{equation}
    \begin{split}
        E(x)=\sigma(W_s(\delta(W_r(x))))
    \end{split}
\end{equation}

\begin{equation}
    \begin{split}
        \sigma(x)={\frac{1}{1+e^{-x}}}
    \end{split}
\end{equation}

In Eq (10), $W_r$ is the fully connected function for compression, and $W_s$ is the fully connected function for reconstruction, which is first downscaled and then upscaled, and then activated by the sSigmoid function to obtain the corresponding weights. The final output is obtained by multiplying the input channels with their respective weights, as in equation (12):

\begin{equation}
    \begin{split}
        \overline{X_x}=F_{scale}(U_c,S_c)=U_c\cdot S_c
    \end{split}
\end{equation}

In summary, the SE module trains a fully connected network with selection capability for feature maps by weights $W_r$, and $W_s$ after converting the multichannel $W\times H$ feature maps into compressed feature vectors, outputs a weight vector for each feature map using the ReLU and sigmoid activation functions, and multiplies it with the original feature map to obtain the final output.

\begin{figure*}[!htbp]
    \centering
    \includegraphics[width=\textwidth]{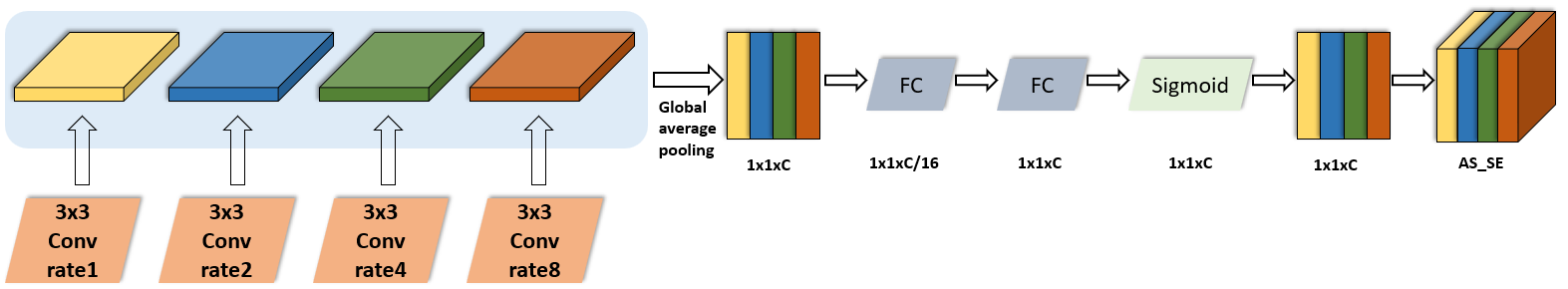}
    \caption{AS-SE model structure}
    \label{AS-SE model structure}
\end{figure*}

\subsubsection{AS-SE Modules}
ASPP comprises multiple atrous convolutions with different sampling rates. The input features were sampled at different sampling rates; that is, the input features were extracted from different scales, and the acquired features were then fused to obtain the final feature extraction results. Compared with SPPF, ASPP can guarantee the resolution while expanding the perceptual field, and the global averaging pooling operation in the ASPP module helps capture the contextual information of the whole image and improve the model's understanding of the scene, which greatly ensures the richness and completeness of feature extraction. In image-processing tasks, we usually want to obtain a large perceptual field while maintaining high resolution. However, these two goals are often contradictory in traditional methods. 
\\\indent
We can use a larger convolution kernel or larger step (stride) to obtain a large perceptual field during the pooling operation. However, both methods have limitations; the former leads to excessive computation, whereas the latter reduces the resolution of the feature map. Dilated convolution is an effective method for solving this contradiction and can expand the perceptual field while keeping the resolution unchanged to extract image features efficiently. This enables a large perceptual field to be obtained without losing excessive resolution. Although the ASPP effectively expands the receptive field by sparsely sampling the input signal, it also leads to a reduction in the correlation between the information obtained from long-distance convolution. To solve this problem, we can use the SENet channel attention mechanism to strengthen the correlation between individual channels and thus better use spatial information in the ASPP. 
\\\indent
Based on the above findings, this paper proposes a module called AS-SE, which incorporates SENet into ASPP. The primary function of the AS-SE module is to multiply the channel weight values generated by SENet with the original input features, which are then used as input features in the atrous convolution, as shown in Fig.16. This combination produces a certain dependency of the information of pavement damage after the atrous convolution on the channel dimension, which in turn enhances the effect of pavement damage feature extraction. By combining SENet with ASPP, the AS-SE module takes full advantage of SENet in terms of channel attention and strengthens the correlation among different channels. This, in turn, aids in extracting more discriminative features. In addition, the AS-SE module maintains the advantages of the ASPP in expanding the field of perception and increasing resolution. Consequently, the overall model performs better in complex pavement damage image-processing tasks.

\subsection{EIoU Loss Function}
The loss function can accurately reflect the difference between the model and the actual data. The loss of border regression in the original YOLOv5 network is calculated using the CIoU loss\cite{zheng2021enhancing} function, which is calculated as follows:

\begin{equation}
    \begin{split}
        \text{L0SS}_{ClOU}=1\text{-IoU}+\frac{\rho^2(b,b^{g t})}{c^2}\text{+$\alpha$V}
    \end{split}
\end{equation}

The IoU\cite{everingham2010pascal} is the ratio of the intersection area of the prediction frame and real frame to the merged area, which is calculated as follows:

\begin{equation}
    \begin{split}
        \text{loU}=\frac{|A\cap B|}{|A\cup B|}
    \end{split}
\end{equation}

Here, $\rho^2(b,b^{g t})$ represents the Euclidean distance between the centroids of the prediction frame and the true frame, and c represents the diagonal distance of the smallest closed region that can contain both the prediction frame and the true frame.

$\alpha$ is the weight function with the following equation:

\begin{equation}
    \begin{split}
        \alpha=\frac{v}{(1-IOU)+v}
    \end{split}
\end{equation}

\textit{v} is a measure of the similarity of the aspect ratios, as defined the following equation:

\begin{equation}
    \begin{split}
        \text{V=}\frac{4}{\pi^2}(arctan\frac{w^{gt}}{h^{gt}}-arctan\frac{w}{h})^2
    \end{split}
\end{equation}

In Eq (16), w, h and $w^{gt}$, $h^{gt}$ represent the predicted frame height and width and the true frame height and width, respectively.
\\\indent 
The CIoU loss covers the overlap area, centroid distance, and aspect ratio of the bounding box regression. However, v in its formula reflects the difference in height and width rather than the true difference between height and width and its confidence level. Therefore, this sometimes prevents the model from effectively optimizing similarity. To address this problem, this improved algorithm uses EIoU loss\cite{zhang2022focal} instead of the original loss function, thus achieving a more efficient loss calculation between the target frame and the prediction frame, which is calculated as follows:

\begin{equation}
    \begin{split}
        LOSS_{EIoU}=1-\text{loU}+\frac{\rho^2(b,b^{gt})}{c^2}+\frac{\rho^2(w,w^{gt})}{c_W^2}+\frac{\rho^2(h,h^{gt})}{c_h^2}
    \end{split}
\end{equation}

In Eq (17), $C_w$ and $C_h$ are the width and height of the minimum outer box covering the target box and the anchor box. The loss function consists of three parts: overlap, center–distance, and width-height losses. The first two parts continue the method in CIoU lLoss, but the width-height loss directly minimizes the difference between the width and height of the target box and anchor box, making the convergence faster.And it makes the regression process more focused on high-quality anchor boxes, which in turn improves the regression accuracy of the predicted boxes; in road disease images, the number of high-quality anchor boxes with small regression errors is usually much smaller than the number of low-quality anchor boxes with large errors, and thus excessive gradients are generated, which affects the results of training, so the introduction of EIoU in this paper also optimizes the sample imbalance in the regression of the edges.

\section{Model training and result analysis}
\subsection{Experimental configuration and parameters}

The experimental environment used in this study was configured with Python Version = 3.7.0, Deep Learning Framework pytorch-gpu = 1.12.1, CUDA Version = 11.6, cuDNN Version = 8.4.0. The parameter settings were Optimizer = SGD, Initial Learning Rate = 0.01, Final learning rate = 0.002, Weight decay factor = 0.0005, Initialization momentum parameter = 0.8, Epoch=300, Batch size = 32.(the best combination of parameters derived from multiple trials)

\subsection{Selection and processing of datasets}
In this study, we collected and collated a total of 12,658 road image datasets. After cleaning a large amount of invalid data, a total of 3644 damage datasets of three categories, namely, 1620 potholes (KC), 1230 cracks (LF), and 794 repairs (XB), and 4000 damage-free datasets, were used for training and recognition of deep learning models according to the actual demand, as shown in Fig.17.

\begin{figure}[H]
    \centering
    \includegraphics[width=0.48\textwidth]{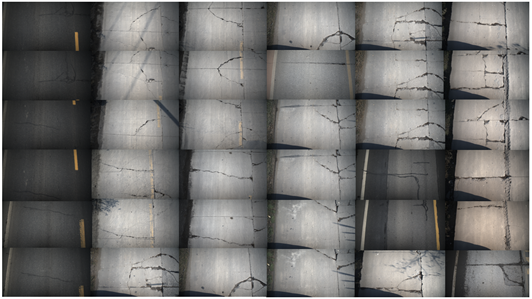}
    \caption{Partial dataset images}
    \label{Partial dataset images}
\end{figure}

In order to enhance the training effect, we balanced the data volume of the three types of damages. After image gradient solving, feature extraction, CycleGAN network generation of diverse backgrounds, and feature and background fusion operations, we expanded the potholes (KC) to 2059, cracks (LF) to 2059, and repairs (XB) to 2058 for a total of 6176 images, and then distributed them into the training set, test set, and validation set according to the 8:1:1 ratio: 4940 sheets for the training set, 618 sheets for the test set, and 618 sheets for the validation set.

\subsection{Evaluation indicators}
In this study, precision, recall, mAP, parameters, floating-point operations per second (flops), and frames per second (FPS) were used to evaluate the performance of the pavement damage detection algorithms.
\\\indent
True Positives(TP):It denotes the number of samples where the resultant information detected by the model exactly matches the information labeled in the training image.
\\\indent
False Negatives(FN):It denotes the number of samples where the resultant information detected by the model does not correspond to the information labeled in the training images.
\\\indent
False Positives(FP):It means that the number of samples which result information detected by the model is correct, while the corresponding training image information is wrong.
\\\indent
True Negatives(TN):It means that the number of samples which information of the results detected by the model is wrong and the information of the training images is also wrong.
\\\indent
Accuracy:The number of positive samples correctly predicted in the algorithmic model as a proportion of the total number of all samples

\begin{equation}
A_{cc}=\begin{matrix}(TP+TN)\\\end{matrix}\Big/_{(TP+FP+FN+TN)}
\end{equation}
\\\indent
Precision:The number of positive samples predicted correctly as a proportion of all instances predicted correctly.

\begin{equation}
P_{re}=^{TP}/_{(TP+FP)}
\end{equation}
\\\indent
Recall:The number of correctly predicted samples as a proportion of the number of all correctly labeled samples.

\begin{equation}
R_{ecall}=^{TP}/_{(TP+FN)}
\end{equation}
\\\indent 
FPS is an evaluation metric of the image-processing speed of the network model when performing detection tasks; the higher the value, the more frames per second can be processed.
\\\indent 
Parameters (params) were used to evaluate the spatial complexity of the model. The number of parameters characterizes the size of the memory occupied by the model and refers to the total number of bytes in each network layer.
\\\indent 
The floating-point operations per second (flops) were used to evaluate the time complexity of the model. The amount of computation characterizes the detection time of the model and refers to the number of floating-point operations per second.

\begin{equation}
\mathrm{FLOPS}=\mathrm{Cores}\times\mathrm{Clock~Speed}\times\mathrm{Operations~Per~Cycle}
\end{equation}

\subsection{Ablation experiment}
To verify the impact of the improved methods in this study on pavement damage detection, the results of the ablation experiments to verify each improved method based on data enhancement are presented in Table 4 (the parameters are derived from the final results of multiple experiments).

\begin{table*}[!t]\large
\tabcolsep=1cm 
\caption{\textbf{Results of ablation experiments}}
\resizebox{\textwidth}{!}{
    \begin{tabular}{@{}lcccccc@{}}
    \toprule
    \textbf{Model} & \textbf{Percision(\%)} & \textbf{Recall(\%)} & \textbf{mAP(\%)} & \textbf{FPS(f/s)} & \textbf{Params(M)} & Flops(G) \\
    \midrule
        YOLOv5&87.1&81.9&85.1&74&7.0&15.8 \\
    
        YOLOv5+CBAM&87.4&82.7&87.1&72&7.1&15.9 \\
    
        YOLOv5+SPPF&88.5&\textcolor{blue}{83.7}&\textcolor{blue}{87.8}&71&\textcolor{blue}{15.2}&\textcolor{blue}{22.4} \\
    
        YOLOv5+EIoU&\textcolor{blue}{89.0}&82.6&87.2&\textcolor{blue}{74}&7.0&15.8 \\
        \textbf{Ours}&\textbf{87.2}&\textbf{85.4}&\textbf{88.2}&\textbf{68}&\textbf{16.1}&\textbf{23.0} \\
    \bottomrule
    \end{tabular}
    }
\end{table*}

The following conclusions can be drawn from Table 4 and Fig.18. After the introduction of the CBAM attention mechanism, the parameters of the whole model and the corresponding computational flops are increased compared with the original YOLOv5, but the detection accuracy mAP value is also improved; after using the EIoU loss to replace the original Loss function CIoU of YOLOv5, the parameters and the computational flops remain unchanged, which improves the detection accuracy at the smallest cost. After using the AS-SE module to replace the original SPPF module, the sensing range is enlarged to make the channel features more obvious, and the high resolution of the image is guaranteed.As shown in Fig.18, when the above improvements were made simultaneously, the effects of the different improvements were not a simple superposition of precision, recall growth, or reduction. Because precision and recall affect each other, increasing precision tends to decrease recall; pursuing an increase in recall usually decreases precision. In the results of the ablation experiments, all improvement methods increased the mAP, which proved the effectiveness of each improvement point.

\begin{figure}[H]
    \centering
    \includegraphics[width=0.48\textwidth]{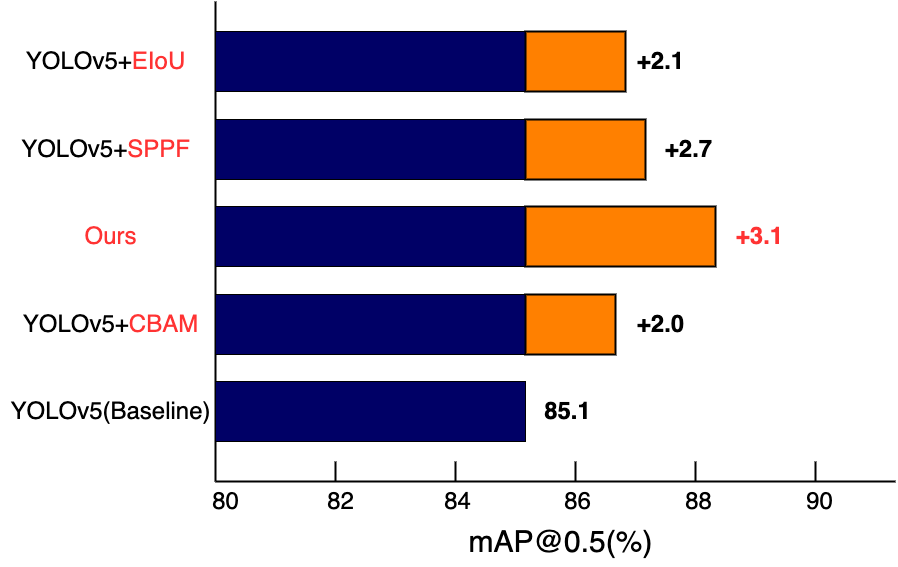}
    \caption{mAP variation of the model under different module combinations}
    \label{mAP variation of the model under different module combinations}
\end{figure}

\subsection{Experimental comparison analysis}

To further evaluate the performance, the proposed algorithms, YOLOv4, YOLOv5, YOLOv7, Faster R-CNN, and SSD target detection algorithm were trained and tested on the dataset of this study. The experimental environment was configured with the above experimental environment to compare and test the performance of different models, and the comparison results are outlined in Table 5, and the dataset test results are shown in Fig.19.

\begin{table*}[!t]
\tabcolsep=1cm 
\caption{\textbf{Experimental results comparison}}
\resizebox{\textwidth}{!}{
    \begin{tabular}{@{}lcccc@{}}
    \toprule
    \textbf{Model} & \textbf{mAP(\%)} & \textbf{FPS(f/s)} & \textbf{Params(M)} & Flops(G) \\
    \midrule

        Faster R-CNN&78.1&22&28.3&948.1 \\

        SSD&72.4&51&26.2&62.7 \\

        YOLOv4&79.2&44&64.3&29.8 \\

        YOLOv5&85.1&\textcolor{blue}{74}&\textcolor{blue}{7.0}&\textcolor{blue}{15.8} \\

        YOLOv7&\textcolor{blue}{87.9}&49&36.5&103.2 \\

        \textbf{Our Model}&\textbf{88.2}&\textbf{68}&\textbf{16.1}&\textbf{23.0} \\
    \bottomrule
    \end{tabular}
    }
\end{table*}

\begin{figure}[H]
    \centering
    \includegraphics[width=0.48\textwidth]{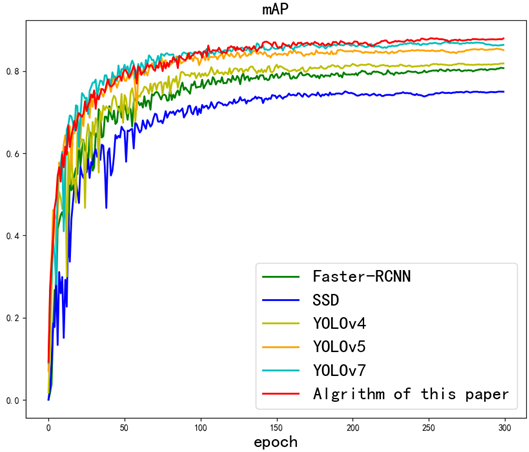}
    \caption{Comparison of mAP change curves of 6 network models}
    \label{Comparison of mAP change curves of 6 network models}
\end{figure}

As presented in Table 5, the Faster R-CNN (ResNet50) algorithm requires the longest detection time; SSD detection is faster, but the detection accuracy is lower. YOLOv7 has the fastest convergence speed and extremely high mAP, but the detection speed is slower than other mainstream target detection algorithms. The improved algorithm in this study on the same dataset achieved an mAP value of 88.2\%, which is the highest among the tested models, 3.1\% higher than the original YOLOv5, and 0.3\% higher than the current more advanced YOLOv7. The lightness of the model is only slightly lower than that of YOLOv5s, and its space complexity and time complexity are much lower than those of YOLOv7. In addition, the network model introduces the AS-SE module, and CBAM module was introduced into the network model, which increased part of the computation, resulting in a decrease in the detection rate of the model compared with the original YOLOv5s model. However, in general, the FPS of the proposed algorithm was as high as 68, which still provided a significant advantage in terms of detection speed.



To verify the detection effectiveness of each of the aforementioned pavement damage detection models on the dataset of this study, three sets of pavement damage feature maps were randomly selected to test the detection effects of YOLOv4, YOLOv5, YOLOv7, Faster R-CNN, SSD, and the improved algorithm. In the labels shown in the detection results, LF represents cracks, KC represents pits, and XB represents repairs. The test results are shown in Fig.20.

\begin{figure}[H]
    \centering
    \includegraphics[width=0.48\textwidth]{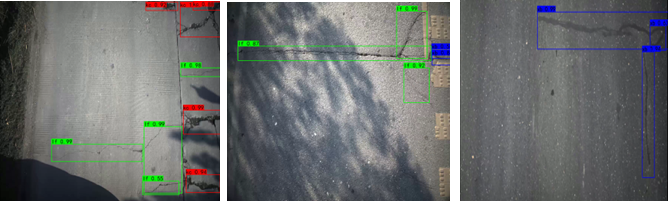}\\
    \centering\textbf{(a).Original Image}
    \label{(a)}
\end{figure}

\vspace{-20pt}
\begin{figure}[H]
    \centering
    \includegraphics[width=0.48\textwidth]{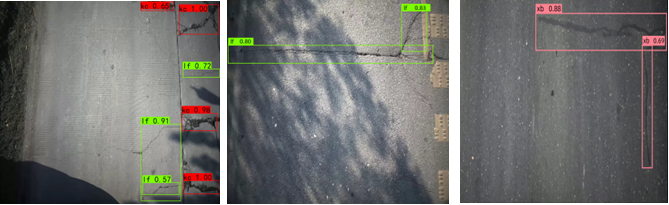}\\
    \centering\textbf{(b).Faster R-CNN}
    \label{(b)}
\end{figure}

\vspace{-20pt}
\begin{figure}[H]
    \centering
    \includegraphics[width=0.48\textwidth]{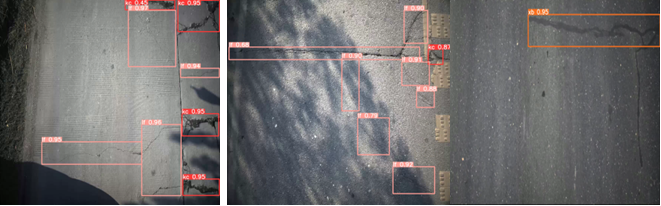}\\
    \centering\textbf{(c).SSD}
    \label{(c)}
\end{figure}

\vspace{-20pt}
\begin{figure}[H]
    \centering
    \includegraphics[width=0.48\textwidth]{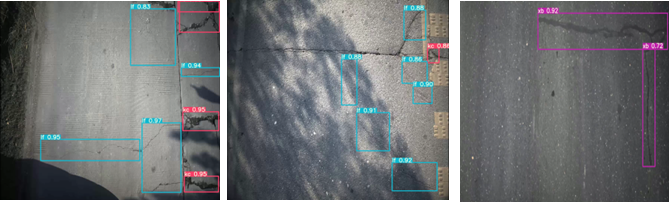}\\
    \centering\textbf{(d).YOLOv4}
    \label{(d)}
\end{figure}

\vspace{-20pt}
\begin{figure}[H]
    \centering
    \includegraphics[width=0.48\textwidth]{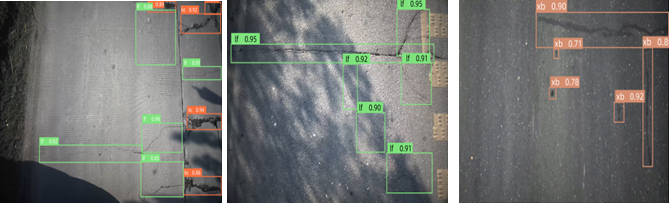}\\
    \centering\textbf{(e).YOLOv5}
    \label{(e)}
\end{figure}

\vspace{-20pt}
\begin{figure}[H]
    \centering
    \includegraphics[width=0.48\textwidth]{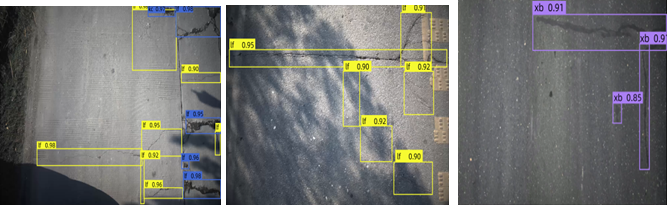}\\
    \centering\textbf{(f).YOLOv7}
    \label{(f)}
\end{figure}

\vspace{-20pt}
\begin{figure}[H]
    \centering
    \includegraphics[width=0.48\textwidth]{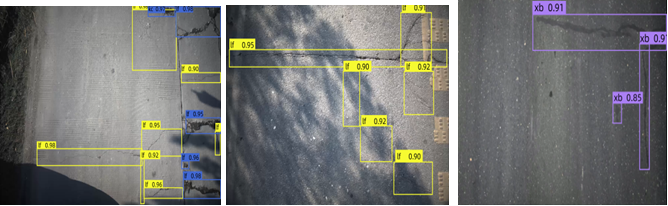}\\
    \centering\textbf{(g).YOLO-ASCE(Ours)}
    \caption{Final test results of 6 models}
    \label{(g)}
\end{figure}

From Fig.20, we can see that the six algorithms can achieve effective detection of pavement damages. Although Faster R-CNN can better identify repairing damages, there are false detections for potholes, which indicates that the algorithm is affected by interference information when extracting features for potholes. This indicates that the algorithm is inaccurate in identifying cracks and is easily influenced by the shadow environment. Because of the small number of low-level feature convolution layers of SSD, the detection of cracks under strong light conditions is missed, which indicates that the shallow feature capability of SSD is not strong enough and the detection accuracy is low. The detection results of YOLOv4 are better for all three types of damages. However, because the target to be detected is closer to the gray value of the background environment, there is a leak in the detection of repair-type damages, and finer repair damages are not detected. The original YOLOv5 missed the detection of cracks and potholes and misidentified cracks as potholes. YOLOv7, as a more advanced target detection algorithm, has higher confidence in all three categories but has detected interfering items as repairs in the repair category. The detection confidence of the improved algorithm was higher for all three categories. The improved algorithm proposed in this paper uses the AS-SE structure, which enhances the feature extraction of the target and introduces the CBAM module, strengthening the target features and enhances the feature expression of the target. It can then accurately detect the target, better addressing the problem of missed detection and maintaining high accuracy even in complex backgrounds.
\\\indent 
The experimental results demonstrate that the proposed YOLOv5 network structure achieves high detection accuracy while maintaining a small variation in detection speed. It emerges as the best choice among the aforementioned network structures in terms of validity and scientificity.

\section{Conclusion}
This study aims to optimize the YOLO-based target detection method to improve the accuracy of pavement damage detection. First, we proposed a data enhancement method based on an improved Scharr filter, CycleGAN, and Laplacian pyramid, which generated image data closer to the real pavement, thus enriching the dataset and indirectly improving the model robustness. Second, we introduced the AS-SE structure to achieve the dependency of the information after atrous convolution in the channel dimension, thus enhancing the feature extraction effect. Next, we employed the CBAM attention mechanism to process the key channel and spatial information in the feature map to focus on this information before feature fusion enhancement. Finally, we applied the EIoU loss function to resist scale variations in YOLOv5 better, reduce localization errors, and improve detection accuracy and generalization capability. Our method demonstrates significantly improved performance through the study and experiments on the pavement damage dataset.
\\\indent
If other techniques can be integrated based on this study to distinguish pavement damages with different damage levels while ensuring detection accuracy and speed, the application prospects of pavement damage detection engineering will be even more promising.

\section*{CRediT auhorship contribution statement}

\textbf{Zhengji Li}:Conceptualization, Methodology, Supervision. \textbf{Xi Xiao}:Conceptualization, Methodology, Validation. \textbf{Jiacheng Xie}:Investigation, Writing - Original Draft, Software. \textbf{Yuxiao Fan}:Formal analysis, Visualization, Software. \textbf{Wentao Wang}:  Writing - Review and Editing. \textbf{Gang Chen}:Data Curation, Resources. \textbf{Liqiang Zhang}:Funding acquisition.

\section*{Declaration of completing interes}

The authors declare that they have no known competing financial interests or personal relationships that could have appeared to influence the work reported in this paper. 

\section*{Data availability}

Data will be made available on request.

\section*{Acknowledgements}

The work is supported by the Project for improving the basic scientific research ability of young and middle-aged teachers in Guangxi Colleges and Universities(NO. 2023KY0884)
\\\indent
We would like to thank Editage(www.editage.cn) for English language editing.

\bibliographystyle{IEEEtran}
\bibliography{IEEEabrv,refrence}

\end{document}